\documentclass[10pt,twocolumn,letterpaper]{article}

\usepackage{cvpr}
\makeatletter
\@namedef{ver@everyshi.sty}{}
\makeatother

\usepackage{diagbox}
\usepackage{tikz}
\usepackage{times}
\usepackage{epsfig}
\usepackage{graphicx}
\usepackage{amsmath}
\usepackage{amssymb}

\usetikzlibrary{arrows}

\usepackage{multirow}
\usepackage{makecell}
\usepackage{xcolor}
\usepackage{enumitem,amssymb}
\usepackage{tkz-kiviat}

\usepackage{pgfplots}
\pgfplotsset{compat=1.14}
\usepackage{pgfplotstable}
\usepgfplotslibrary{statistics}
\usetikzlibrary{pgfplots.groupplots}

\usepackage[export]{adjustbox}
\usepackage[tight,footnotesize]{subfigure}
\usepackage{booktabs}
\DeclareMathOperator{\sign}{sign}
\usepackage[noend]{algorithmic}
\usepackage[algoruled,nofillcomment,algo2e]{algorithm2e}
\usetikzlibrary{spy}
\newcommand{\myparagraph}[1]{\vspace{1ex}\noindent{\bf #1}}

\SetKwComment{Comment}{$\triangleright$\ }{}

\definecolor{blue-violet}{rgb}{0.54, 0.17, 0.89}

\DeclareMathOperator*{\argmin}{arg\,min}

\definecolor{darkgreen}{RGB}{47,109,79}
\definecolor{darkblue}{RGB}{57,79,99}
\definecolor{rosso}{RGB}{220,57,18}
\definecolor{giallo}{RGB}{255,153,0}
\definecolor{blu}{RGB}{102,140,217}
\definecolor{verde}{RGB}{16,150,24}
\definecolor{viola}{RGB}{153,0,153}
\definecolor{awesome}{rgb}{1.0, 0.13, 0.32}
\definecolor{ref}{rgb}{0.65,0.65,0.65} 
\definecolor{darkgreen}{RGB}{47,109,79}
\definecolor{darkblue}{RGB}{57,79,99}
\definecolor{rosso}{RGB}{220,57,18}
\definecolor{verde}{RGB}{16,150,24}
\definecolor{viola}{RGB}{153,0,153}
\definecolor{americanrose}{rgb}{1.0, 0.01, 0.24}
\definecolor{bostonuniversityred}{rgb}{0.8, 0.0, 0.0}
\definecolor{chocolate(traditional)}{rgb}{0.48, 0.25, 0.0}
\definecolor{violet(web)}{rgb}{0.93, 0.51, 0.93}
\definecolor{airforceblue}{rgb}{0.36, 0.54, 0.66}

\definecolor{almond}{rgb}{0.94, 0.87, 0.8}
\definecolor{amethyst}{rgb}{0.6, 0.4, 0.8}
\definecolor{bazaar}{rgb}{0.6, 0.47, 0.48}
\definecolor{britishracinggreen}{rgb}{0.0, 0.26, 0.15}
\definecolor{byzantine}{rgb}{0.74, 0.2, 0.64}
\definecolor{cadetblue}{rgb}{0.37, 0.62, 0.63}
\definecolor{cambridgeblue}{rgb}{0.64, 0.76, 0.68}
\definecolor{candypink}{rgb}{0.89, 0.44, 0.48}
\definecolor{caputmortuum}{rgb}{0.35, 0.15, 0.13}
\definecolor{cerulean}{rgb}{0.0, 0.48, 0.65}
\definecolor{corn}{rgb}{0.98, 0.93, 0.36}
\definecolor{darkbyzantium}{rgb}{0.36, 0.22, 0.33}
\definecolor{darkgoldenrod}{rgb}{0.72, 0.53, 0.04}
\definecolor{darkseagreen}{rgb}{0.56, 0.74, 0.56}
\definecolor{darkturquoise}{rgb}{0.0, 0.81, 0.82}
\definecolor{dimgray}{rgb}{0.41, 0.41, 0.41}
\definecolor{eggplant}{rgb}{0.38, 0.25, 0.32}

\definecolor{R50}{rgb}{0.0, 0.81, 0.82}
\definecolor{R18}{rgb}{0.41, 0.41, 0.41}
\definecolor{AN}{rgb}{0.38, 0.25, 0.32}

\definecolor{detColBit}{rgb}{0,0,1}
\definecolor{detColMedian}{rgb}{0.25,0.41,1.0}
\definecolor{detColJPEG}{rgb}{0.29,0,0.51}

\definecolor{BR}{RGB}{175,0,3}
\definecolor{MR}{RGB}{233,107,34}
\definecolor{JR}{RGB}{38,181,66}

\usepackage{placeins}
\usepackage{etoolbox}
\usepackage{colortbl}
\newrobustcmd{\B}{\bfseries}
\usepackage{empheq}
\definecolor{blue-violet}{rgb}{0.54, 0.17, 0.89}
\newcommand{\unaryminus}{\scalebox{0.55}[1.0]{\( - \)}}

\usepackage[pagebackref=true,breaklinks=true,letterpaper=true,colorlinks,bookmarks=false]{hyperref}
\usepackage{import}
\cvprfinalcopy 

\def\cvprPaperID{7606}

\ifcvprfinal\fi
\begin{document}

\title{ColorFool: Semantic Adversarial Colorization}

\author{Ali Shahin Shamsabadi, Ricardo Sanchez-Matilla,  Andrea Cavallaro\\
Centre for Intelligent Sensing\\
Queen Mary University of London, UK\\
{\tt\small \{a.shahinshamsabadi,ricardo.sanchezmatilla,a.cavallaro\}@qmul.ac.uk}
}

\maketitle

\begin{abstract}
Adversarial attacks that generate small $L_p$-norm perturbations to mislead classifiers have limited  success in black-box settings and with unseen classifiers. These attacks are also not robust to defenses that use denoising filters and to adversarial training procedures. Instead, adversarial attacks that generate unrestricted perturbations are more robust to defenses, are generally more successful in black-box settings and are more transferable to unseen classifiers. However, unrestricted perturbations may be noticeable to humans. In this paper, we propose a content-based black-box adversarial attack that generates unrestricted perturbations by exploiting image semantics to selectively modify colors within chosen ranges that are perceived as natural by humans. We show that the proposed approach, ColorFool, outperforms in terms of success rate, robustness to defense frameworks and transferability, five state-of-the-art adversarial attacks on two different tasks, scene and object classification, when attacking three state-of-the-art deep neural networks using three standard datasets. The source code is available at {\url{https://github.com/smartcameras/ColorFool}}.
\end{abstract}

\section{Introduction}

Adversarial attacks perturb the intensity values of  a {\em  clean} image to mislead machine learning classifiers, such as Deep Neural Networks (DNNs). These perturbations can be restricted~\cite{dong2019evading,kurakin2016adversarial,Li2019,modas2018sparsefool,MoosaviDezfooli16,papernot2016limitations} or {unrestricted}~\cite{bhattad2019big,hosseini2018semantic} with respect to the intensity values in the clean image. Restricted perturbations, which are generated by controlling an $L_p$-norm, may restrain the maximum change for each pixel ($L_\infty$-norm~\cite{dong2019evading,kurakin2016adversarial,Li2019}), the maximum number of perturbed pixels ($L_0$-norm~\cite{modas2018sparsefool,papernot2016limitations}), or the maximum energy change ($L_2$-norm~\cite{MoosaviDezfooli16}); whereas unrestricted perturbations span a wider range, as determined by different colorization approaches~\cite{bhattad2019big,hosseini2018semantic}.

Defenses against adversarial attacks apply  re-quantization~\cite{xu2017feature}, median filtering~\cite{xu2017feature} and JPEG compression~\cite{das2017keeping,dziugaite2016study} to remove adversarial perturbations prior to classification, or improve the robustness of the classifier through adversarial training~\cite{goodfellow2014explaining}, or by changing the loss functions~\cite{Mustafa_2019_ICCV}. The property of {\em robustness} of an adversarial attack is the success rate of misleading a classifier in the presence of defense frameworks.  Most adversarial attacks assume white-box settings, i.e.~the attacker has full knowledge of the  architecture and parameters (and hence gradients) of the classifier~\cite{dong2019evading,kurakin2016adversarial,Li2019,modas2018sparsefool,MoosaviDezfooli16,papernot2016limitations}. However, real-world scenarios may prevent access to the classifier (unseen classifier) or limit the exposed information to only the output of the classifier (black-box settings). The property of {\em transferability} is the success rate of adversarial images in misleading an unseen classifier~\cite{demontis2019adversarial}. Finally, the perturbation in an adversarial image should be {\em unnoticeable}, i.e.~the shape and spatial arrangement of objects in the adversarial image should be perceived as in the clean image and the colors should look natural.

\begin{figure}[t!]
    \centering
    \setlength\tabcolsep{0.5pt}
    \begin{tabular}{cc}
    \begin{tikzpicture}
        \node[inner sep=0pt] (russell) at (0,0) {\includegraphics[width=0.4\columnwidth]{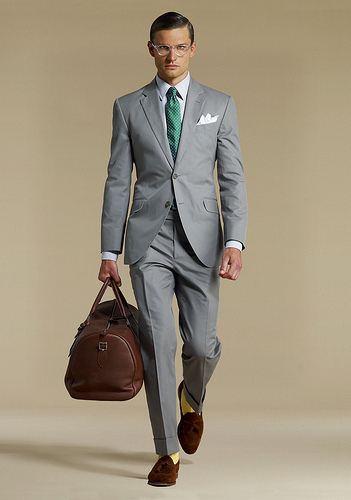}};
        \draw[fill=black] (27pt,-60.5pt) rectangle (47pt,-67.5pt) node[pos=.5] {\tiny \textcolor{white}{suit}};
    \end{tikzpicture}&
    \begin{tikzpicture}
        \node[inner sep=0pt] (russell) at (0,0) {\includegraphics[width=0.4\columnwidth]{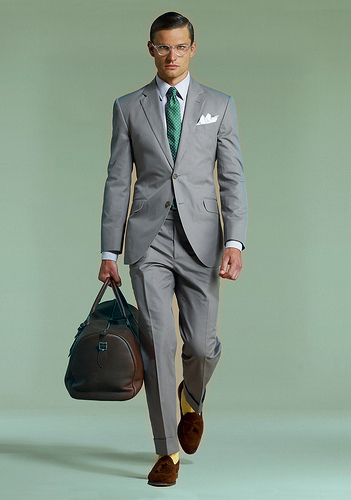}};
        \draw[fill=black] (27pt,-60.5pt) rectangle (47pt,-67.5pt) node[pos=.5] {\tiny \textcolor{white}{racket}};
    \end{tikzpicture}\\
    (a) & (b) \\
    \end{tabular}\\
    \begin{tabular}{ccccc}
        \begin{tikzpicture}
            \node[inner sep=0pt] (russell) at (0,0) {\includegraphics[width=0.18\columnwidth]{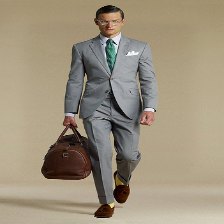}};
            \draw[fill=black] (-1pt,-14pt) rectangle (21pt,-21pt) node[pos=.5] {\tiny \textcolor{white}{kimono}};
        \end{tikzpicture}&
        \begin{tikzpicture}
            \node[inner sep=0pt] (russell) at (0,0) {\includegraphics[width=0.18\columnwidth]{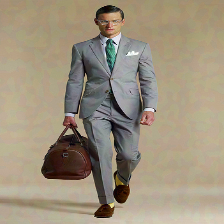}};
            \draw[fill=black] (-1pt,-14pt) rectangle (21pt,-21pt) node[pos=.5] {\tiny \textcolor{white}{racket}};
        \end{tikzpicture}&
        \begin{tikzpicture}
            \node[inner sep=0pt] (russell) at (0,0) {\includegraphics[width=0.18\columnwidth]{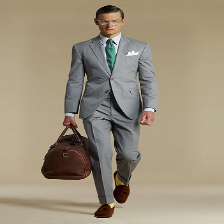}};
            \draw[fill=black] (-1pt,-14pt) rectangle (21pt,-21pt) node[pos=.5] {\tiny \textcolor{white}{kimono}};
        \end{tikzpicture}&
        \begin{tikzpicture}
            \node[inner sep=0pt] (russell) at (0,0) {\includegraphics[width=0.18\columnwidth]{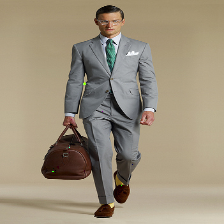}};
            \draw[fill=black] (-1pt,-14pt) rectangle (21pt,-21pt) node[pos=.5] {\tiny \textcolor{white}{loafer}};
        \end{tikzpicture}&
        \begin{tikzpicture}
            \node[inner sep=0pt] (russell) at (0,0) {\includegraphics[width=0.18\columnwidth]{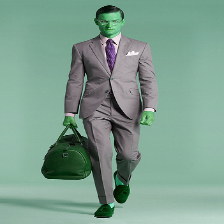}};
            \draw[fill=black] (-1pt,-14pt) rectangle (21pt,-21pt) node[pos=.5] {\tiny \textcolor{white}{trench coat}};
        \end{tikzpicture}\\
        (c) & (d) & (e) & (f) & (g) 
    \end{tabular}
      \caption{Adversarial image generated for a sample (a) clean image by (b) ColorFool, (c) Basic Iterative Method (BIM)~\cite{kurakin2016adversarial}, (d) Translation-Invariant BIM~\cite{dong2019evading}, (e) DeepFool~\cite{MoosaviDezfooli16}, (f) SparseFool~\cite{modas2018sparsefool} and (g) SemanticAdv~\cite{hosseini2018semantic}. BIM and DeepFool generate unnoticeable adversarial images with restricted perturbations. ColorFool generates any-size, natural-color adversarial images by considering the semantic information and preserving the colors of regions within an image that the human vision system is more sensitive to (in this case the person). The text in the bottom right of each image indicates the predicted class.
      }
      \label{fig:OrgSizeAdv}
\end{figure}

Restricted perturbations~\cite{carlini2017towards,kurakin2016adversarial,Li2019,MoosaviDezfooli16} often have high spatial frequencies that can be detected by defenses~\cite{das2017keeping,dziugaite2016study,liu2018feature,xu2017feature}. Moreover, restricted perturbations that are sparse and with large changes in intensity are noticeable~\cite{modas2018sparsefool,papernot2016limitations}. 
Instead, unrestricted attacks arbitrarily perturb intensity values through a colorization process~\cite{bhattad2019big}, which is based on an expensive training phase, followed by per-image adversarial fine-tuning. Alternatively, attacks can arbitrarily change the hue and saturation components in the $HSV$ color space~\cite{hosseini2018semantic}. However, even small variations can result in large and perceivable distortions caused by unnatural colors (see Fig.~\ref{fig:OrgSizeAdv}(g)).

In this paper, we propose a black-box, unrestricted, content-based adversarial attack that exploits the characteristics of the human visual system to selectively alter colors. The proposed approach, \emph{ColorFool}, operates only on the de-correlated $a$ and $b$ channels of the perceptually uniform $Lab$ color space~\cite{ruderman1998statistics}, without changing the lightness, $L$. Moreover, ColorFool introduces perturbations only within a chosen natural-color range for specific semantic categories~\cite{zhang2016colorful}. Unlike other adversarial attacks, the proposed adversarial perturbation can be generated for images of any size (see Fig.~\ref{fig:OrgSizeAdv}(b)).  We validate ColorFool in attacking three state-of-the-art DNNs (ResNet50, ResNet18~\cite{he2016deep} and AlexNet~\cite{krizhevsky2012imagenet}) that have been trained for scene and object classification tasks using three datasets (ImageNet~\cite{deng2009imagenet}, CIFAR-10~\cite{krizhevsky2009learning} and Private Places365 (P-Places365)~\cite{zhou2017places}). We show that ColorFool generates natural-color adversarial images that are effective in terms of success rate in seen and unseen classifiers, robustness to defense frameworks on an extensive comparison with five state-of-the-art attacks.

\section{Adversarial attacks}
\label{sec:relatedWork}
\begin{table*}[t!]
\centering
\caption{Comparison of adversarial attacks. An adversarial image is generated with a  perturbation restricted by ${L_0}$, ${L_1}$, ${L_2}$ and ${L_\infty}$ or unrestricted perturbation on the \textbf{C}olors considering two \textbf{Type}s: \textbf{W}hite- or \textbf{B}lack-attack for two \textbf{Task}s: \textbf{O}bject and \textbf{S}cene classification. \textbf{Datasets} with the number of chosen classes for the attack are reported for ImageNet, CIFAR-10 and Private-Places365 (P-Places365) with 1000, 10 and 60 classes as well (\textbf{U}nknown, if number of classes are not written in their papers). JSMA is tested on the MNIST dataset (10 classes). KEY-- BIM: Basic Iterative Method; TI-BIM: Translation-Invariant BIM; P-BIM: Private BIM; CW: Carlini-Wagner; JSMA: Jacobian-based Saliency Map Attack; and SemAdv: Semantic Adversarial.}
\resizebox{\textwidth}{!}{
\begin{tabular}{rlcclcccc}
\Xhline{3\arrayrulewidth}
\multirow{2}{*}{Ref} & \multirow{2}{*}{Attack} & \multirow{2}{*}{Perturbation} & \multirow{2}{*}{Type} & \multirow{2}{*}{Attacked classifier} & \multicolumn{3}{c}{Datasets} & \multirow{2}{*}{Task} \\
&  &  &  &   & {ImageNet} & {CIFAR-10} & {P-Places365} & \\
\hline
\cite{kurakin2016adversarial} & BIM & $L_\infty$  & W & Inc-v3  & 1000 & & & O \\
\cite{dong2019evading} & TI-BIM (``TI-FGSM") & $L_\infty$  & W & Inc-v3, Inc-v4, ResNet152  & 60 & &  & O\\
\cite{Li2019}& P-BIM (``P-FGSM") & $L_\infty$  & W & ResNet50  && & 60 & S\\
\cite{MoosaviDezfooli16} & DeepFool & $L_2$  & W & LeNet, CaffeNet, GoogleNet   & 1000 & 10&  & O\\
\cite{modas2018sparsefool} & SparseFool & $L_1$  & W & LeNet, ResNet18, Inc-v3, DenseNet,VGG16   & U &10& & O\\
\cite{papernot2016limitations} & JSMA &$L_0$  & W & LeNet  & & && O \\
\cite{carlini2017towards} & CW & $L_{0,2,\infty}$ & W & Inc-v3  & 1000 & 10 &  & O\\
\cite{bhattad2019big} & BigAdv & C & W  & ResNet50, DenseNet121, VGG19  & 10 & &  & O\\
\cite{hosseini2018semantic} & SemAdv & C & B & VGG16   & &10 & & O\\
ours & ColorFool & C & B & ResNet50, ResNet18, AlexNet  & 1000 & 10 & 60 & O and S\\
\Xhline{3\arrayrulewidth}
\end{tabular}
}
\label{tab:RelatedWork}
\end{table*}
Let $\mathbf{X} \in {\mathbb{Z}}^{w,h,c}$ be an $RGB$ clean image with width $w$, height $h$ and $c=3$ color channels. Let $M(\cdot)$ be a DNN classifier that predicts for a given image the most probable class, $y= M(\mathbf{X})$. An adversarial attack perturbs $\mathbf{X}$ to generate an adversarial image, $\mathbf{\dot{X}}$, such that {$M(\mathbf{\dot{X}}) \neq M(\mathbf{X})$}.

Adversarial attacks can be grouped based on their perturbations into two categories, namely restricted and unrestricted. An adversarial image can be generated with a perturbation controlled  by $L_0$, $L_1$, $L_2$ or $L_\infty$-norms. Adversarial attacks that use restricted perturbations are Basic Iterative Method {(BIM)}~\cite{kurakin2016adversarial}, Translation-Invariant BIM {(TI-BIM)}~\cite{dong2019evading}, {DeepFool}~\cite{MoosaviDezfooli16} and {SparseFool}~\cite{modas2018sparsefool}. Alternatively, an adversarial image can be generated with an unrestricted perturbation on the colors considering the preservation of the shape of the objects, as in {SemanticAdv}~\cite{hosseini2018semantic} or {BigAdv}~\cite{bhattad2019big}.

{BIM}~\cite{kurakin2016adversarial} constrains the maximum perturbation of each pixel by imposing an $L_\infty$-norm constraint. BIM searches for adversarial images by linearising the cost function, $J_M(\cdot)$, in the input space. The search starts from $\dot{\mathbf{X}}_0 = \mathbf{X}$ and iteratively moves in the direction of the gradient of the cost of predicting $y$ with respect to the input image, $\nabla_{\mathbf{X}}J_M(\cdot)$, with step size $\delta$ in each iteration: 
\begin{equation}
\label{eq:fgsm_iter}
      \dot{\mathbf{X}}_N = C_{\mathbf{X},\epsilon} \bigg(\dot{\mathbf{X}}_{N-1} + \delta \sign\Big(\nabla_{\mathbf{X}}J_M\big(\mathbb{\theta},\dot{\mathbf{X}}_{N-1},y\big)\Big)\bigg),
\end{equation}
until $M(\dot{\mathbf{X}}_N) \neq M({\mathbf{X}})$ or a maximum number of $N$ iterations, where $\theta$ are the parameters of $M(\cdot)$, $\sign(\cdot)$ is the sign function that determines the direction of the gradient of the cost function. $C_{\mathbf{X},\epsilon}(\cdot)$ is a clipping function that maintains the adversarial images within the $\epsilon$-neighborhood of the clean image as well as $[0,255]$:
\begin{equation}
    C_{\mathbf{X},\epsilon} (\dot{\mathbf{X}}) = \text{min} \left\{\mathbf{255}, \mathbf{X} + \epsilon, \text{max}\left\{ \mathbf{0}, \mathbf{X}-\epsilon,\dot{\mathbf{X}}\right\} \right\},
\end{equation}
where $\mathbf{0}$ and $\mathbf{255}$ are images whose pixel intensities are all $0$ and $255$, respectively, and $\text{min}(\cdot)$/$\text{max}(\cdot)$ are the per-pixel min/max operation.

{TI-BIM}~\cite{dong2019evading} generates BIM adversarial perturbations over an ensemble of translated images to improve the transferability to unseen classifiers. As the gradients of a translated image correspond to translating the gradient of the original image~\cite{dong2019evading}, TI-BIM convolves the gradient with a pre-defined kernel $\mathbf{W}$ at each iteration:
\begin{equation}
\label{eq:Ti-BIM}
      \dot{\mathbf{X}}_N = C_{\mathbf{X},\epsilon} \bigg(\dot{\mathbf{X}}_{N-1} + \delta \sign\Big(\mathbf{W}*\nabla_{\mathbf{X}}J_M\big(\mathbb{\theta},\dot{\mathbf{X}}_{N-1},y\big)\Big)\bigg),
\end{equation}
where $\mathbf{W}$ can be a uniform, linear or Gaussian kernel. 

{DeepFool}~\cite{MoosaviDezfooli16} finds the minimal $L_2$-norm adversarial perturbation by finding the direction towards the closest decision boundary.
For example in the case of the binary classifier, adversarial images can iteratively be generated by projecting the adversarial image of each iteration onto the closest linearized decision boundary of $M(\cdot)$: 
\begin{equation}
    \label{eq:DF}
    \dot{\mathbf{X}}_N = \mathbf{X} + (1+\eta)\sum_{n=1}^{N}-\frac{M(\dot{\mathbf{X}}_n)}{\|\nabla M(\dot{\mathbf{X}}_n)\|_2^2}\nabla M(\dot{\mathbf{X}}_n),
\end{equation}
where $\eta \ll 1$ is a constant that is multiplied by the accumulative adversarial perturbations to reach the other side of the decision boundary. Note that DeepFool does not impose constraints on pixel values, which, as a result, may lie outside the permissible dynamic range. 

{SparseFool}~\cite{modas2018sparsefool} uses the $L_1$-norm between the clean and adversarial images to minimize the number of perturbed pixels. SparseFool leverages the fact that DNNs have a low mean curvature in the neighborhood of each image~\cite{fawzi2018empirical} and generates sparse perturbations  based on this curvature and adversarial images on the closest $L_2$ decision boundary. SparseFool approximates the decision boundary near the clean image $\mathbf{X}$ by a hyperplane, $\mathbf{v}^T \dot{\mathbf{X}}_{\mbox{\scriptsize DF}}$, passing the minimal $L_2$-norm DeepFool adversarial image (i.e. Eq~\ref{eq:DF}), $\dot{\mathbf{X}}_{\mbox{\scriptsize DF}}$, and a normal vector $\mathbf{v}$. Then, SparseFool iteratively finds the minimal $L_1$-norm projection of the clean image onto the approximated decision boundary:
\begin{equation}
    \dot{\mathbf{X}}_N = D(\dot{\mathbf{X}}_{N-1} + \boldsymbol{\delta^*}),
\end{equation}
where $\boldsymbol{\delta^*}$ is the sparse adversarial perturbation and $D(\cdot)$ is a clipping function that maintains the pixel values between $[0,255]$: 

\begin{equation}
    D(\dot{\mathbf{X}}) = \text{min} \left\{\mathbf{255}, \text{max}\{ \mathbf{0},\dot{\mathbf{X}} \} \right\}.
\end{equation}
Each $d$-th value of SparseFool perturbation, $\delta^*_d$, is iteratively computed as
\begin{equation}
    \delta^*_d = \frac{|\mathbf{v}^T(\dot{\mathbf{X}}_n - \dot{\mathbf{X}}_{\mbox{\scriptsize DF}})|}{|v_d|}\sign(v_d),
\end{equation}
where $T$ is the transpose operator. 

{SemanticAdv}~\cite{hosseini2018semantic} unrestrictedly changes colors in the $HSV$ color space by shifting the hue, $\mathbf{X}_H$, and saturation, $\mathbf{X}_S$, of the clean image while preserving the value channel, $\mathbf{X}_V$, in order to not affect the shapes of objects: 
\begin{align}
    \dot{{\mathbf{X}}}_N = \beta(\begin{bmatrix}
           \mathbf{X}_H+[\mathbf{\delta}_H]^{w,h},
           \mathbf{X}_S + [\mathbf{\delta}_S]^{w,h},
           \mathbf{X}_V
         \end{bmatrix}),
\label{eq:SemAdv}         
\end{align}
where $\mathbf{\delta}_S$, $\mathbf{\delta}_H$ $\in [0,1]$ are scalar random values, and $\beta(\cdot)$ is a function that converts the intensities from the $HSV$ to the $RGB$ color space. Eq.~\ref{eq:SemAdv} is repeated until ${M(\dot{\mathbf{X}}_N) \neq M(\mathbf{X})}$ or a maximum number of trials (1000) is reached. 

{BigAdv}~\cite{bhattad2019big} aims to generate  natural-color perturbations by fine-tuning, for each $\mathbf{X}$, a trained colorization model~\cite{zhang2016colorful}, $F(\cdot)$, parameterized by $\hat{\theta}$ with a cross-entropy adversarial loss $J_{adv}$~\cite{carlini2017towards}:
\begin{equation}
    \dot{\mathbf{X}}=\argmin_{\hat{\theta}}J_{adv}(M\big(F(\mathbf{X}_L,\mathbf{C}_{h},\mathbf{L}_{h}; \hat{\theta})\big),y),
\end{equation}
where $\mathbf{X}_L$ is the $L$ value of the image in the $Lab$ color space and $\mathbf{C}_h$ is the ground-truth color for the locations that are indicated by the binary location hint, $\mathbf{L}_h$. BigAdv de-colorizes the whole image and again colorizes it. This process may severely distort the colors if $\mathbf{C}_h$ and $\mathbf{L}_h$ are not carefully set.

Finally, ColorFool, the proposed approach (see Sec.~\ref{sec:colorfool}), is an unrestricted, black-box attack like SemanticAdv. However, SemanticAdv perturbs pixel intensities without considering the content in an image thus often producing unnatural colors. ColorFool instead perturbs colors only in specific semantic regions and within a chosen range so they can  be still perceived as natural. The other state-of-the-art unrestricted attack, BigAdv, is a white-box attack that trains a colorization model to learn image statistics from large datasets and to fine-tune the model for each image. Tab.~\ref{tab:RelatedWork} summarizes the adversarial attacks for object or scene classification tasks.

\section{ColorFool}
\label{sec:colorfool}

We aim to design a black-box adversarial attack that generates  adversarial images with natural colors through generating low-frequency perturbations that are highly transferable to unseen classifiers and robust to defenses. Moreover, the attack shall operate on the native size of the images. 

First, we identify image regions whose color is important for a human observer as the appearance of these sensitive regions (e.g. human skin) is typically within a specific range. Other (non-sensitive) image regions (e.g. wall and curtains in Figs.~\ref{fig:OrgSizeAdv}~and~\ref{fig:advExamples}, first row), instead, may have their colors modified within an arbitrary range and still look natural~\cite{charpiat2008automatic}. We consider four categories of sensitive regions, whose unusual colors would attract the attention of a human observer~\cite{charpiat2008automatic,levin2004colorization,zhang2016colorful}: person, sky, vegetation (e.g.~grass and trees), and water (e.g.~sea, river, waterfall, swimming pool and lake).  

\begin{figure}
    \centering
    \setlength\tabcolsep{1pt}
    \begin{tabular}{cccc}
         \includegraphics[width=0.24\columnwidth]{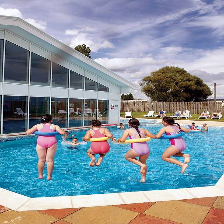}&
         \includegraphics[width=0.24\columnwidth]{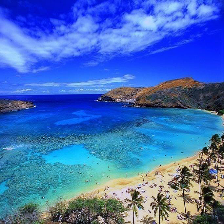}&
         \includegraphics[width=0.24\columnwidth]{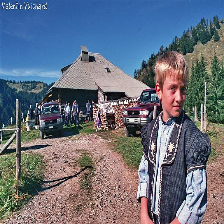}&
         \includegraphics[width=0.24\columnwidth]{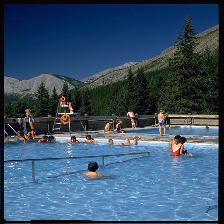} \\
         \includegraphics[width=0.24\columnwidth]{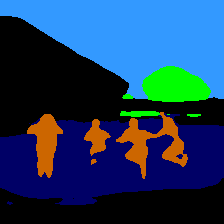}&
         \includegraphics[width=0.24\columnwidth]{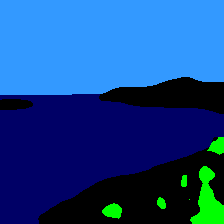}&
         \includegraphics[width=0.24\columnwidth]{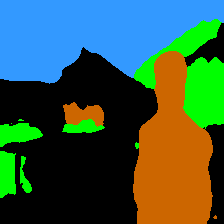}&
         \includegraphics[width=0.24\columnwidth]{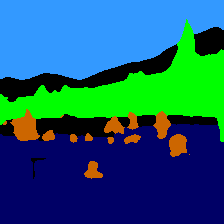} \\
    \end{tabular}
    \caption{Sample results of image semantic segmentation~\cite{zhou2017scene}. 
    ColorFool  identifies non-sensitive regions (in black) and color-sensitive semantic regions, namely person (in orange), vegetation (in green), sky (in light blue) and water (in dark blue).}
    \label{fig:AllMasks}
\end{figure}

Let us decompose an image $\mathbf{X}$ into $K$ semantic regions
\begin{equation}
    \mathcal{S} = \left\{ \mathbf{S}_k : \mathbf{S}_k = \mathbf{X} \cdot  \mathbf{M}_k \right\}_{k=1}^{K},
\end{equation}
where $\mathbf{M}_k \in \{0,1\}^{w,h}$ is a binary mask that specifies the location of pixels belonging to region $\mathbf{S}_k$ and ``$\cdot$'' denotes a pixel-wise multiplication. Binary masks are outputted by a pyramid Pooling R50-Dilated architecture of Cascade Segmentation Module segmentation~\cite{zhou2018semantic},  trained on the MIT ADE20K dataset~\cite{zhou2017scene} on 150 semantic region types. Fig.~\ref{fig:AllMasks} shows examples of the considered semantic regions.

We separate the sensitive regions, $\mathbb{S} = \{ \mathbf{S}_k \}_{k=1}^{S}$, from the non-sensitive regions, $\overline{\mathbb{S}} = \{ \overline{\mathbf{S}}_k \}_{k=1}^{\overline{S}}$, where $\mathcal{S} = \mathbb{S} \cup \overline{\mathbb{S}}$ and $\cup$ is the union operator. 
After identifying these two sets, we appropriately modify the colors of the region in the perceptually uniform $Lab$ color space~\cite{ruderman1998statistics}, which  separates color information from brightness: $a$ ranges from green (-128) to red (+127), $b$ ranges from blue (-128) to yellow (+127), and $L$ ranges from black (0) to white (100). 

We then modify the color of the sensitive regions, $\mathbb{S}$, to generate the adversarial set $ \dot{\mathbb{S}}$ as
\begin{equation}
    \label{eq:CFattack}
    \dot{\mathbb{S}} = \{ \dot{\mathbf{S}}_k : \dot{\mathbf{S}}_k = \gamma(\mathbf{S}_k) + \alpha {[0,  N^a_k, N^b_k]}^T \}_{k=1}^{S},
\end{equation}
where $\gamma(\cdot)$ converts the intensities of an image from the $RGB$ to the $Lab$ color space, $N^a_k \in \mathcal{N}_k^a$ and $N^b_k \in \mathcal{N}_k^b$ are the adversarial perturbations in the channels $a$ and $b$ that are chosen randomly from the set of natural-color ranges~\cite{zhang2016colorful}, $\mathcal{N}_k^a$ and $\mathcal{N}_k^b$, in the $a$ and $b$ channels. These ranges are defined based on the actual colors, region semantics and prior knowledge about color perception in that region type (see Tab.~\ref{tab:AdvColor}). 
We allow multiple trials, until a perturbation misleads the classifier. Let $n$ be the index of the trial and $N$ be the maximum number of trials. To avoid large color changes in the first trials, we progressively scale the randomly chosen perturbation by $\alpha=\frac{n}{N}$.

\begin{table}[t!]
\centering
\caption{Adversarial color perturbation considered by ColorFool to modify the colors of sensitive semantic regions. The natural-color ranges are chosen based on the color recommendation of people to gray-scale objects~\cite{charpiat2008automatic,levin2004colorization} that are also used as ground-truth colors in colorization methods~\cite{zhang2016colorful}. The adversarial color perturbation of the $k$-th semantic region considers the extreme values of the semantic class as $l_k^a = \text{min}(\mathbf{S}_k)$ and $u_k^a = \text{max}(\mathbf{S}_k)$. The adversarial perturbation is chosen randomly within each natural-color range and applied as in Eq.~\ref{eq:CFattack}. Note that no color changes are applied to image regions classified as person.}
\vspace{0.2cm}
\resizebox{\columnwidth}{!}{
\begin{tabular}{lll}
    \Xhline{3\arrayrulewidth}
    {Semantic region} & \multicolumn{1} {c}{{$a$ channel}} & \multicolumn{1} {c}{{$b$ channel}} \\  
    \cmidrule(lr){1-1}
    \cmidrule(lr){2-2}
    \cmidrule(lr){3-3}
    $\mathcal{S}_1$: Person & $\mathcal{N}^a_1 =\{0\}$ & $\mathcal{N}^b_1=\{0\}$ \\
    $\mathcal{S}_2$: Vegetation & $\mathcal{N}^a_2 =\{\unaryminus 128 \unaryminus l^a_2,\dots,\unaryminus u^a_2\}$ & $\mathcal{N}^b_2=\{\unaryminus l^b_2,\dots,127 \unaryminus u^b_2\}$ \\
    $\mathcal{S}_3$: Water & $\mathcal{N}^a_3 =\{\unaryminus 128 \unaryminus l^a_3,\dots,\unaryminus u^a_3\}$ & $\mathcal{N}^b_3=\{\unaryminus 128 \unaryminus l^b_3,\dots,\unaryminus u^b_3\}$ \\
    $\mathcal{S}_4$: Sky & $\mathcal{N}^a_4 =\{\unaryminus 128 \unaryminus l^a_4,\dots, \unaryminus u^a_4\}$ & $\mathcal{N}^b_4=\{\unaryminus 128 \unaryminus l^b_4,\dots, \unaryminus u^b_4\}$  \\
    \Xhline{3\arrayrulewidth}
\end{tabular}
}
\label{tab:AdvColor}
\end{table}

We modify the color of the non-sensitive regions, $\overline{\mathbb{S}}$, to produce the set $\dot{\overline{\mathbb{S}}}$ as
\begin{equation}
    \dot{\overline{\mathbb{S}}} = \{ \dot{\overline{\mathbf{S}}}_k : \dot{\overline{\mathbf{S}}}_k = \gamma(\overline{\mathbf{S}}_k) + \alpha{[0,\overline{N}^a,\overline{N}^b]}^T \}_{k=1}^{\overline{S}},
\end{equation}
where $\overline{N}^a \in \{-127, \dots, 128\}$ and $\overline{N}^b \in \{-127, \dots, 128\}$ are chosen randomly inside the whole range of $a$ and $b$, as the regions can undergo larger intensity changes. 

Finally, the adversarial image $\dot{\mathbf{X}}$ generated by ColorFool combines the modified sensitive and non-sensitive image regions as
\begin{equation}
    \dot{\mathbf{X}} = Q \bigg( \gamma^{-1} \Big(
        \sum_{k=1}^{S}{\dot{\mathbf{S}}_k}+\sum_{k=1}^{\overline{S}}{\dot{\overline{\mathbf{S}}}_k}
     \Big) \bigg),
\end{equation}
where $Q(\cdot)$ is the quantization function which ensures that the generated adversarial image is in the dynamic range of the pixel values, $\dot{\mathbf{X}} \in {\mathbb{Z}}^{w,h,c}$, and $\gamma^{-1}(\cdot)$ is the inverse function that converts the intensities of an image from the $Lab$ to the $RGB$ color space.

\section{Validation}
\label{sec:validation}

\myparagraph{Algorithms under comparison}.
We compare the proposed attack, ColorFool, against the state-of-the-art adversarial attacks discussed in Section~\ref{sec:relatedWork}: Basic Iterative Method (BIM)~\cite{kurakin2016adversarial}, Translation-Invariant BIM (TI-BIM)~\cite{dong2019evading}, DeepFool~\cite{MoosaviDezfooli16}, SparseFool~\cite{modas2018sparsefool} and SemanticAdv~\cite{hosseini2018semantic} (we excluded BigAdv~\cite{bhattad2019big} as no code was available at the time of submission). These attacks include restricted and unrestricted perturbations and generate adversarial images that are transferable (TI-BIM), unnoticeable (DeepFool) and robust to defenses (SemanticAdv). We also compare against the simple yet successful BIM attack and SparseFool, a sparse attack.
Furthermore, we consider a modification of the proposed attack, named {ColorFool-r}, where no priors are considered for the semantic regions. We use the authors' implementations for all adversarial attacks apart from SemanticAdv that we re-implemented in PyTorch. All adversarial images are generated using the same read/write framework, image filters and software version in PyTorch and OpenCV to make the results comparable.

\myparagraph{Datasets.} 
We use three datasets Private-Places365 (P-Places365)~\cite{zhou2017places}, a scene classification dataset; CIFAR-10~\cite{krizhevsky2009learning}, an object classification dataset; and ImageNet~\cite{deng2009imagenet}, another object classification dataset.
For P-Places365, we employ a subset of classes that were defined as sensitive in the MediaEval 2018 Pixel Privacy Challenge~\cite{Mediaeval2018}. P-Places365 includes 50 images for each of the 60 private scene classes.
For CIFAR-10, we use the whole test set, which is composed of 10K images of 10 different object classes.
For ImageNet, we consider the 1000 classes and 3 random images per class from the  validation set. 
All the images are $RGB$ with varying resolution except for the images from CIFAR-10 whose $w=h=32$.

\myparagraph{Classifiers under attack}.
We conduct the attacks on two different architectures: a deep residual neural network (ResNet~\cite{he2016deep}, 18 layers (R18) and 50 layers (R50)) and AlexNet (AN)~\cite{krizhevsky2012imagenet}. 
We choose these three classifiers to study the transferability comparing both homogeneous (i.e.~ResNet classifiers) and heterogeneous architectures (i.e.~AlexNet).

\myparagraph{Performance measures.}
We quantify the \emph{success rate} in misleading a classifier, the \emph{robustness} to defenses and the \emph{image quality} of the adversarial images. 
The \emph{success rate} (SR) is quantified as the ratio between the number of adversarial images that mislead the classifier on its most-likely predicted class and the total number of images. For the transferability, we compute the SR of adversarial images generated for a seen classifier in misleading unseen classifiers.
The \emph{robustness} to defenses is measured as follows. Firstly, we quantify the SR in seen classifiers of adversarial images after filtering. As filters we use re-quantization~\cite{xu2017feature} with 1 to 7 bits, in steps of 1; median filtering~\cite{xu2017feature}, with squared kernel of dimension 2, 3 and 5; and lossy JPEG compression~\cite{dziugaite2016study,das2017keeping}, with quality parameters 25, 50, 75 and 100. We report the results on retrieving the class that was predicted on the clean images with the most effective filter (i.e. the one that obtains the lowest SR).
Secondly, we report the undetectability as the ratio between adversarial images not identified as adversarials and the total number of images using the previously mentioned image filters~\cite{xu2017feature}. Specifically, for each classifier and parameter of each image filter, we compute a threshold that determines if an image is adversarial or clean by comparing the $L_1$-norm of the difference between the prediction probability vectors of the given image and the same image after the image filtering. Each threshold is calculated as the value that allows for a $5\%$ false-positive rate in detecting clean images on a training dataset. Then, images with $L_1$-norm difference larger than the threshold are considered to be adversarials.
Thirdly, we evaluate the SR when attacking a seen classifier trained with Prototype Conformity Loss (PCL)~\cite{Mustafa_2019_ICCV} and adversarial training~\cite{goodfellow2014explaining}. 
Finally, we quantify the image quality of the adversarial image with a non-reference perceptual image quality measure named neural image assessment (NIMA)~\cite{Esfandarani2017nima} trained on the AVA dataset~\cite{Murray2012}. NIMA estimates the perceived image quality and was shown to predict human preferences~\cite{Mediaeval2018}.

\begin{table}[t]
    \centering
    \setlength{\tabcolsep}{1.3pt}
    \caption{Success rate on Private-Places365 (P-Places365), CIFAR-10 and ImageNet datasets against ResNet50 (R50), ResNet18 (R18) and AlexNet (AN). The performance of these classifiers on the clean images is presented in the third row. The higher the success rate, the most successful the attack. KEY-- AC: attacked classifier; TC: test classifier; Acc: accuracy. A gray (white) cell denotes a seen (unseen) classifier.
    ColorFool is more transferable than other adversarial attacks, except SemanticAdv, which however severely distort the colors of all regions (see Fig.~\ref{fig:advExamples}).}
    \vspace{0.2cm}
    \resizebox{\columnwidth}{!}{
    \begin{tabular}{l|c|ccc|ccc|ccc}
        \Xhline{3\arrayrulewidth}
        \multirow{2}{*}{Attack} & {Dataset} & \multicolumn{3}{c|}{P-Places365} &  \multicolumn{3}{c|}{CIFAR-10} & \multicolumn{3}{c}{ImageNet} \\ \cline{2-11}
        & \diagbox[width=5em]{AC}{TC} & {R50} & {R18} & {AN} & {R50} & {R18} & {AN} & {R50} & {R18} & {AN} \\
        \Xhline{3\arrayrulewidth} 
        \multicolumn{2}{c|}{Acc. on \emph{clean} images} & .554 & .527 & .466 & .944 & .935 & .722 & .726 & .649 & .517  \\
        \Xhline{3\arrayrulewidth} 
        \multirow{3}{*}{{BIM}} & {R50} & 
        \cellcolor[HTML]{EFEFEF}1.00  & .284 & .073 & \cellcolor[HTML]{EFEFEF}.999 & .095 & .021 & \cellcolor[HTML]{EFEFEF}.873 &  .123 & .087 \\
         & {R18} & .231  & \cellcolor[HTML]{EFEFEF}1.00 & .081 & .078 & \cellcolor[HTML]{EFEFEF}.999 & .022 & .143  & \cellcolor[HTML]{EFEFEF}.945 & .099 \\
         & {AN} & .061  & .081 & \cellcolor[HTML]{EFEFEF}1.00 & .014 & .013 & \cellcolor[HTML]{EFEFEF}.999 & .088  & .092 & \cellcolor[HTML]{EFEFEF}.944 \\
         \hline
         \multirow{3}{*}{{TI-BIM}} & {R50} & \cellcolor[HTML]{EFEFEF}.995 & .339 & .186  & \cellcolor[HTML]{EFEFEF}.843 & .153 & .173 & \cellcolor[HTML]{EFEFEF}.992 & .235 & .176 \\
         & {R18} & .268 & \cellcolor[HTML]{EFEFEF}.996 & .198  & .083 & \cellcolor[HTML]{EFEFEF}.943 & .138 & .173 & \cellcolor[HTML]{EFEFEF}.997 & .183 \\
         & {AN} & .157 & .193 & \cellcolor[HTML]{EFEFEF}.995  & .315 & .349 & \cellcolor[HTML]{EFEFEF}.889 & .121 & .163 & \cellcolor[HTML]{EFEFEF}.994 \\
         \hline
        \multirow{3}{*}{{DF}} & {R50} & \cellcolor[HTML]{EFEFEF}.957 & .107  & .030  & \cellcolor[HTML]{EFEFEF}.829 & .226 & .064 & \cellcolor[HTML]{EFEFEF}.983 & .071 & .018 \\
        & {R18} & .009  & \cellcolor[HTML]{EFEFEF}.969  & .030 & .234 & \cellcolor[HTML]{EFEFEF}.875 & .076 & .055 & \cellcolor[HTML]{EFEFEF}.991 & .017 \\
        & {AN} & .021  & .028 & \cellcolor[HTML]{EFEFEF}.956  & .020 & .024 &  \cellcolor[HTML]{EFEFEF}.637  & .017 & .019 & \cellcolor[HTML]{EFEFEF}.993 \\
         \hline
        \multirow{3}{*}{{SF}} & {R50} & \cellcolor[HTML]{EFEFEF}.998 & .151  & .127  & \cellcolor[HTML]{EFEFEF}.999 & .408 & .186 & \cellcolor[HTML]{EFEFEF}.987 & .167 & .176 \\
        & {R18} & .101 & \cellcolor[HTML]{EFEFEF}.999  & .120  & .353  & \cellcolor[HTML]{EFEFEF}.999 & .216 & .086 & \cellcolor[HTML]{EFEFEF}.997 & .134 \\
        & {AN} & .070 & .066 & \cellcolor[HTML]{EFEFEF}1.00  & .130 & .151 &  \cellcolor[HTML]{EFEFEF}.999 & .062 & .079 & \cellcolor[HTML]{EFEFEF}.999 \\
         \hline
        \multirow{3}{*}{{SA}} & {R50} & \cellcolor[HTML]{EFEFEF}.936  & .563 & .713 & \cellcolor[HTML]{EFEFEF}.863 & .429 &.704 & \cellcolor[HTML]{EFEFEF}.889 & .540  & .769 \\
        & {R18} & .480  & \cellcolor[HTML]{EFEFEF}.954 & .714 & .339 & \cellcolor[HTML]{EFEFEF}.898 &  .705 & .422 & \cellcolor[HTML]{EFEFEF}.931  & .757 \\
        & {AN} & .424  & .466 & \cellcolor[HTML]{EFEFEF}.990   & .155 & .191 & \cellcolor[HTML]{EFEFEF}.993 & .359 & .431  & \cellcolor[HTML]{EFEFEF}.994 \\
         \hline
        \multirow{3}{*}{{CF-r}} & {R50} & \cellcolor[HTML]{EFEFEF}.963  & .336 & .514 & \cellcolor[HTML]{EFEFEF}.956 & .255 & .635 & \cellcolor[HTML]{EFEFEF}.948 & .362 & .608 \\
        & {R18} & .275  & \cellcolor[HTML]{EFEFEF}.970 & .501 & .431 & \cellcolor[HTML]{EFEFEF}.954 & .689 & .235 & \cellcolor[HTML]{EFEFEF}.954 & .580 \\
        & {AN} & .157  & .171 & \cellcolor[HTML]{EFEFEF}.999 & .065 & .058 & \cellcolor[HTML]{EFEFEF}.999 & .104 & .137 & \cellcolor[HTML]{EFEFEF}.998 \\
         \hline
        \multirow{3}{*}{{CF}} & {R50} &  \cellcolor[HTML]{EFEFEF}.959   & .334 & .491 & \cellcolor[HTML]{EFEFEF}.975 & .254 & .641 & \cellcolor[HTML]{EFEFEF}.917 & .348 & .592 \\
         & {R18} &  .267   & \cellcolor[HTML]{EFEFEF}.971 & .475 & .415 & \cellcolor[HTML]{EFEFEF}.971 & .696 & .223 & \cellcolor[HTML]{EFEFEF}.934 & .543 \\
        & {AN} & .171   & .157 & \cellcolor[HTML]{EFEFEF}.998 & .059 & .055 &  \cellcolor[HTML]{EFEFEF}1.00 & .114 & .147 & \cellcolor[HTML]{EFEFEF}.995 \\
        \Xhline{3\arrayrulewidth}
    \end{tabular}
    }
    \label{tab:TR-PC}
\end{table}
\myparagraph{Success rate}.
Tab.~\ref{tab:TR-PC} shows the SR on a seen classifier (on-diagonal elements) and transferability to unseen classifiers (off-diagonal elements).
All adversarial attacks achieve high SR in a seen classifier for most of the classifiers and datasets. Restricted attacks never achieve SRs higher than 0.41 in unseen classifiers, while unrestricted attacks achieve a SR of up to 0.77.
ColorFool achieves a high SR on both seen and unseen classifiers with, for example, 0.97 when both attacking and testing in R18 in CIFAR-10 and 0.69 and 0.41 when evaluated with AN and R50, respectively. However, other attacks only achieve SRs of 0.02 (BIM), 0.14 (TI-BIM), 0.07 (DeepFool), 0.21 (SparseFool). A possible reason is that restricted attacks such as BIM, iteratively overfit to the parameters of the specific classifier, which means that the adversarial images rarely mislead other classifiers, while the randomness in changing the color in ColorFool prevents this overfitting. TI-BIM overcomes the overfitting of BIM and achieves higher transferability than BIM, while its SR in seen classifiers decreases. 
Unrestricted attacks obtain high transferability rates. For instance, in the CIFAR-10 dataset, SemanticAdv, {ColorFool-r} and ColorFool obtain SRs of 0.71, 0.69 and 0.70 when attacking R18 and evaluating in AN. While ColorFool outperforms SemanticAdv with seen classifiers, SemanticAdv obtains higher transferability rates. This is due to the large color changes that SemanticAdv introduces in the whole image, including regions that are more informative for the classifier (higher transferability) but also regions that are sensitive for the human vision system, thus generating unnatural colors (see Fig.~\ref{fig:advExamples}). Further insights are discussed in the image quality analysis later in this section.
As previously studied~\cite{Liu2016}, adversarial images generated on stronger classifiers (e.g. R50) have a higher transferability rate when tested on weaker classifiers (e.g. AN). This behavior can be observed, for instance, when looking at the results of ColorFool in P-Places365. Adversarial images crafted with R50 obtain a SR of $0.96$, which decreases to $0.49$ when tested in AN. However, when adversarial images are crafted with AN the SR is $0.99$, but when tested in R50 (a stronger classifier) the SR is only $0.17$.

\pgfplotsset{ every non boxed x axis/.append style={x axis line style=-}}
\begin{figure*}[t]
\centering

\caption{Robustness of Basic Iterative Method (BIM), Translation-Invariant BIM (TI-B), DeepFool (DF), SparseFool (SF), SemanticAdv (SA), ColorFool-r (CF-r) and ColorFool (CF) on ResNet50, ResNet18 and AlexNet against re-quantization~({\protect\tikz \protect\draw[color=BR, fill=BR] plot[mark=*, mark size=0.5mm] (0,0);}), median filtering~({\protect\tikz \protect\draw[color=MR, fill=MR] plot[mark=*,mark size=0.5mm] (0,0);}) and JPEG compression~({\protect\tikz \protect\draw[color=JR, fill=JR] plot[mark=*, mark size=0.5mm] (0,0);}) on the Private subset of Places365 (P-Places365), CIFAR-10 and ImageNet.}
\label{fig:robustness}
\end{figure*}

\myparagraph{Robustness to defenses.} 
The SR of adversarial attacks after applying any of the three image filters is depicted in Fig.~\ref{fig:robustness}. Restricted attacks such as DeepFool and SparseFool are the least robust to image filtering, as these filters can remove restricted adversarial noises (especially $L_0$ sparse adversarial perturbation) prior to the classification and correctly classify around $70\%$ of them. BIM and TI-BIM obtain higher SR than other restricted attacks in P-Places365 and ImageNet but similar in CIFAR-10. The most robust attacks are the unrestricted ones where SemanticAdv, ColorFool-r and ColorFool consistently obtain a SR above $60\%$ across datasets and classifiers.
The undetectability results (Fig.~\ref{fig:detectability}) show that restricted attacks are more detectable than unrestricted ones when considering all image filters across all classifiers and datasets.
For instance, when attacking R50 in P-Places365, BIM, TI-BIM, DeepFool and SparseFool obtain undetectability rates of 5\%, 19\%, 1\% and 11\% with re-quantization, median filtering and JPEG compression.
Unrestricted attacks such as SemanticAdv, ColorFool-r and ColorFool obtain 73\%, 72\% and 75\%.
We believe that one reason for this is related to the spatial frequency of the generated adversarial perturbations. Restricted attacks generate high-frequency adversarial perturbations, whereas unrestricted attacks generate low-frequency perturbations (see Fig.~\ref{fig:advExamples}).
Low-frequency perturbations (those generated by unrestricted attacks) are more robust to re-quantization, median filtering and JPEG compression. In general, JPEG compression is the most effective detection framework.
\begin{table}[t]
    \centering
    \caption{Success rate of Basic Iterative Method (BIM), Translation-Invariant BIM (TI-BIM), DeepFool (DF), SparseFool (SF), SemanticAdv (SA), ColorFool-r (CF-r) and ColorFool (CF) against ResNet110 trained with softmax, on Prototype Conformity Loss (PCL)~\cite{Mustafa_2019_ICCV} and its combination with adversarial training (AdvT)~\cite{goodfellow2014explaining} on CIFAR-10. The higher the success rate, the more robust the attack. In bold, the best performing attack.}
    \vspace{0.2cm}
    \resizebox{\columnwidth}{!}{
    \begin{tabular}{lccccccc}
    \Xhline{3\arrayrulewidth} 
         \multirow{2}{*}{Training}&\multirow{1}{*}{{BIM}} & \multirow{1}{*}{{TI-BIM}} & \multicolumn{1}{c}{{DF}} & \multicolumn{1}{c}{{SF}} & \multicolumn{1}{c}{{SA}}& \multicolumn{1}{c}{{CF-r}} & \multicolumn{1}{c}{{CF}}  \\
                  &\cite{kurakin2016adversarial} & \cite{dong2019evading} &\cite{MoosaviDezfooli16}& \cite{modas2018sparsefool} & \cite{hosseini2018semantic}& ours & ours  \\
         
         \hline
         {Softmax}& .969 & .963 & .855 & \textbf{.994} & .867 & .992 & \textbf{.994}\\
         {PCL}& .560 & .619 & .784 & .801 & .896 & \textbf{1.00} & \textbf{1.00}\\
         {PCL+AdvT}& .500 & .577 & .665 & .691 & .966 & .998 & \textbf{.999}\\
    \Xhline{3\arrayrulewidth} 
    \end{tabular}
    }
    \label{tab:R-Adv-PCL}
\end{table}
When we consider all of the filters applied, as an example, to P-Places365, the restricted attacks BIM, TI-BIM, DeepFool and SparseFool are detectable in 95\%, 81\%, 99\%	 and 89\% of the cases. However, unrestricted attacks such as SemanticAdv and ColorFool-r are detectable in only 27\% of the cases and ColorFool is the least detectable (25\%).

\definecolor{FGc}{RGB}{         0,  113.9850,  188.9550}
\definecolor{TIFGc}{RGB}{216.7500,   82.8750,   24.9900}
\definecolor{DFc}{RGB}{236.8950,  176.9700,  31.8750}
\definecolor{SFc}{RGB}{131.8966  131.8966  255.0000}
\definecolor{SAc}{RGB}{0  153  0}
\definecolor{CFrc}{RGB}{0  204  0}
\definecolor{CFc}{RGB}{0  255  0}

\pgfplotstableread{places_points.txt}\detectPlaces
\pgfplotstableread{cifar_points.txt}\detectCIFAR
\pgfplotstableread{imagenet_points.txt}\detectImageNet
\begin{figure}[t!]
    \centering
    \scriptsize
    \begin{tikzpicture}
    \begin{axis}[
        width=\columnwidth,
        xmin=0,xmax=64,
        ymin=0, ymax=1,
        ytick={0,0.5,1.0},
        ylabel={Undetect.},
        ylabel near ticks,
        y=15mm,
        xticklabels={,,},
        xtick=data,
        ymajorgrids=true,
        tick label style={font=\scriptsize},
        label style={font=\scriptsize},
        xtick style={draw=none},
        ]
    \addplot+[only marks, mark=square*, mark size=1pt, mark options={color=FGc}] table[y expr=1-\thisrow{Fgy},x=Fgx]{\detectPlaces};
    \addplot+[only marks, mark=square*, mark size=1pt, mark options={color=TIFGc}] table[y expr=1-\thisrow{TIFGy},x=TIFGx]{\detectPlaces};
    \addplot+[only marks, mark=square*, mark size=1pt, mark options={color=DFc}] table[y expr=1-\thisrow{Dfy},x=Dfx]{\detectPlaces};
    \addplot+[only marks, mark=square*, mark size=1pt, mark options={color=SFc}] table[y expr=1-\thisrow{Sfy},x=Sfx]{\detectPlaces};
    \addplot+[only marks, mark=square*, mark size=1pt, mark options={color=SAc}] table[y expr=1-\thisrow{Say},x=Sax]{\detectPlaces};
    \addplot+[only marks, mark=square*, mark size=1pt, mark options={color=CFrc}] table[y expr=1-\thisrow{Cfry},x=Cfrx]{\detectPlaces};
    \addplot+[only marks, mark=square*, mark size=1pt, mark options={color=CFc}] table[y expr=1-\thisrow{Cfy},x=Cfx]{\detectPlaces};
    \end{axis}
    \begin{axis}[
        width=\columnwidth,
        xmin=0,xmax=64,
        ymin=0, ymax=1,
        y=15mm,
        ylabel={P-Places365},
        ytick={0,0.5,1.0},
        yticklabels={},
        ylabel near ticks,
        yticklabel pos=right,
        tick label style={font=\scriptsize},
        label style={font=\scriptsize},
        xticklabels={,,},
        xtick style={draw=none},
        ]
    \end{axis}
    \end{tikzpicture}
    \\
    \begin{tikzpicture}
    \begin{axis}[
        width=\columnwidth,
        xmin=0,xmax=64,
        ymin=0, ymax=1,
        ytick={0,0.5,1.0},
        ylabel={Undetect.},
        ylabel near ticks,
        y=15mm,
        xticklabels={,,},
        xtick=data,
        ymajorgrids=true,
        tick label style={font=\scriptsize},
        label style={font=\scriptsize},
        xtick style={draw=none},
        ]
     \addplot+[only marks, mark=square*, mark size=1pt, mark options={color=FGc}] table[y expr=1-\thisrow{Fgy},x=Fgx]{\detectCIFAR};
    \addplot+[only marks, mark=square*, mark size=1pt, mark options={color=TIFGc}] table[y expr=1-\thisrow{TIFGy},x=TIFGx]{\detectCIFAR};
    \addplot+[only marks, mark=square*, mark size=1pt, mark options={color=DFc}] table[y expr=1-\thisrow{Dfy},x=Dfx]{\detectCIFAR};
    \addplot+[only marks, mark=square*, mark size=1pt, mark options={color=SFc}] table[y expr=1-\thisrow{Sfy},x=Sfx]{\detectCIFAR};
    \addplot+[only marks, mark=square*, mark size=1pt, mark options={color=SAc}] table[y expr=1-\thisrow{Say},x=Sax]{\detectCIFAR};
    \addplot+[only marks, mark=square*, mark size=1pt, mark options={color=CFrc}] table[y expr=1-\thisrow{Cfry},x=Cfrx]{\detectCIFAR};
    \addplot+[only marks, mark=square*, mark size=1pt, mark options={color=CFc}] table[y expr=1-\thisrow{Cfy},x=Cfx]{\detectCIFAR};
    \end{axis}
    \begin{axis}[
        width=\columnwidth,
        xmin=0,xmax=64,
        ymin=0, ymax=1,
        y=15mm,
        ylabel={CIFAR-10},
        ytick={0,0.5,1.0},
        yticklabels={},
        ylabel near ticks,
        yticklabel pos=right,
        tick label style={font=\scriptsize},
        label style={font=\scriptsize},
        xticklabels={,,},
        xtick style={draw=none},
        ]
    \end{axis}
    \end{tikzpicture}\\
    \begin{tikzpicture}
    \begin{axis}[
        width=\columnwidth,
        xmin=0,xmax=64,
        ymin=0, ymax=1,
        ytick={0,0.5,1.0},
        ylabel={Undetect.},
        ylabel near ticks,
        y=15mm,
        ytick={0,0.5,1.0},
        xticklabels={\emph{Re-quantization}, \emph{Median}, \emph{JPEG}},
        xtick={12,33,54},
        ymajorgrids=true,
        tick label style={font=\scriptsize},
        label style={font=\scriptsize},
        xtick style={draw=none},
        ]
    \addplot+[only marks, mark=square*, mark size=1pt, mark options={color=FGc}] table[y expr=1-\thisrow{Fgy},x=Fgx]{\detectImageNet};
    \addplot+[only marks, mark=square*, mark size=1pt, mark options={color=TIFGc}] table[y expr=1-\thisrow{TIFGy},x=TIFGx]{\detectImageNet};
    \addplot+[only marks, mark=square*, mark size=1pt, mark options={color=DFc}] table[y expr=1-\thisrow{Dfy},x=Dfx]{\detectImageNet};
    \addplot+[only marks, mark=square*, mark size=1pt, mark options={color=SFc}] table[y expr=1-\thisrow{Sfy},x=Sfx]{\detectImageNet};
    \addplot+[only marks, mark=square*, mark size=1pt, mark options={color=SAc}] table[y expr=1-\thisrow{Say},x=Sax]{\detectImageNet};
    \addplot+[only marks, mark=square*, mark size=1pt, mark options={color=CFrc}] table[y expr=1-\thisrow{Cfry},x=Cfrx]{\detectImageNet};
    \addplot+[only marks, mark=square*, mark size=1pt, mark options={color=CFc}] table[y expr=1-\thisrow{Cfy},x=Cfx]{\detectImageNet};
    \end{axis}
    \begin{axis}[
        width=\columnwidth,
        xmin=0,xmax=64,
        ymin=0, ymax=1,
        y=15mm,
        ylabel={ImageNet},
        ytick={0,0.5,1.0},
        yticklabels={},
        ylabel near ticks,
        yticklabel pos=right,
        tick label style={font=\scriptsize},
        label style={font=\scriptsize},
        xticklabels={,,},
        xtick style={draw=none},
        ]
    \end{axis}
    \end{tikzpicture}
  \caption{Undetectability (Undetec.) of
   \protect\tikz \protect\draw[FGc,fill=FGc] (0,0) rectangle (1.ex,1.ex);~Basic Iterative Method (BIM),
   \protect\tikz \protect\draw[TIFGc,fill=TIFGc] (0,0) rectangle (1.ex,1.ex);~Translation-Invariant BIM (TI-BIM),
   \protect\tikz \protect\draw[DFc,fill=DFc] (0,0) rectangle (1.ex,1.ex);~DeepFool,
   \protect\tikz \protect\draw[SFc,fill=SFc] (0,0) rectangle (1.ex,1.ex);~SparseFool,
   \protect\tikz \protect\draw[SAc,fill=SAc] (0,0) rectangle (1.ex,1.ex);~SemanticAdv,
   \protect\tikz \protect\draw[CFrc,fill=CFrc] (0,0) rectangle (1.ex,1.ex);~ColorFool-r and
   \protect\tikz \protect\draw[CFc,fill=CFc] (0,0) rectangle (1.ex,1.ex);~ColorFool, when attacking ResNet50, ResNet18 and AlexNet classifiers (first, second and third square of each color, respectively) using re-quantization, median filtering and JPEG compression. The higher the undetectability, the higher the robustness to defenses.
   }
  \label{fig:detectability}
\end{figure}

\begin{figure*}[t!]
    \centering
    \setlength\tabcolsep{0pt}
    \resizebox{\textwidth}{!}{
    \begin{tabular}{cccccccc}
         Clean & BIM~\cite{kurakin2016adversarial} & TI-BIM~\cite{dong2019evading} & DF~\cite{MoosaviDezfooli16} & SF~\cite{modas2018sparsefool} & SA~\cite{hosseini2018semantic} & CF-r & CF \\
         \adjustbox{raise=5.45ex}{
        \begin{tikzpicture}
            \node[inner sep=0pt] (russell) at (0,0) {\includegraphics[width=0.12\textwidth]{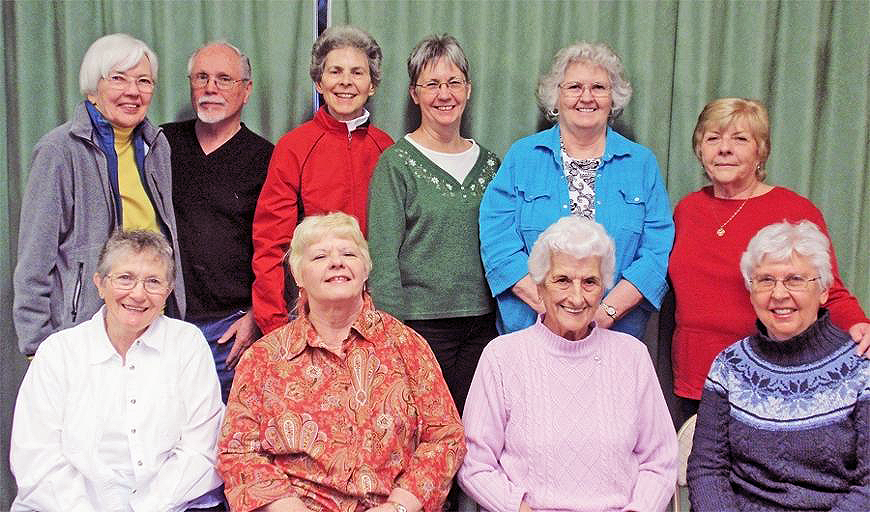}};
            \draw[fill=black] (-0.5pt,-11pt) rectangle (29.5pt,-18pt) node[pos=.5] {\tiny \textcolor{white}{nursing room}};
        \end{tikzpicture}
         }&
         \begin{tikzpicture}
            \node[inner sep=0pt] (russell) at (0,0) {\includegraphics[width=0.12\textwidth]{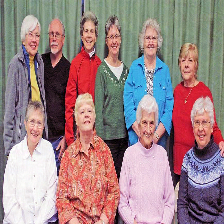}};
            \draw[fill=black] (-0.5pt,-23pt) rectangle (29.5pt,-30pt) node[pos=.5] {\tiny \textcolor{white}{butchers shop}};
        \end{tikzpicture}&
         \begin{tikzpicture}
            \node[inner sep=0pt] (russell) at (0,0) {\includegraphics[width=0.12\textwidth]{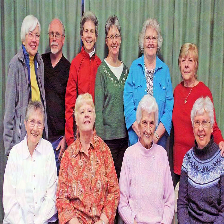}};
            \draw[fill=black] (-0.5pt,-23pt) rectangle (29.5pt,-30pt) node[pos=.5] {\tiny \textcolor{white}{butchers shop}};
        \end{tikzpicture}&
        \begin{tikzpicture}
            \node[inner sep=0pt] (russell) at (0,0) {\includegraphics[width=0.12\textwidth]{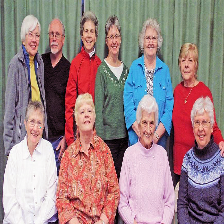}};
            \draw[fill=black] (-0.5pt,-23pt) rectangle (29.5pt,-30pt) node[pos=.5] {\tiny \textcolor{white}{army base}};
        \end{tikzpicture}&
         \begin{tikzpicture}
            \node[inner sep=0pt] (russell) at (0,0) {\includegraphics[width=0.12\textwidth]{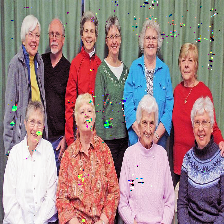}};
            \draw[fill=black] (-0.5pt,-23pt) rectangle (29.5pt,-30pt) node[pos=.5] {\tiny \textcolor{white}{army base}};
        \end{tikzpicture}&
        \begin{tikzpicture}
            \node[inner sep=0pt] (russell) at (0,0) {\includegraphics[width=0.12\textwidth]{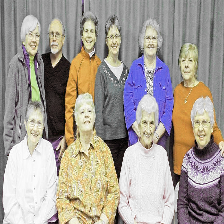}};
            \draw[fill=black] (-0.5pt,-23pt) rectangle (29.5pt,-30pt) node[pos=.5] {\tiny \textcolor{white}{florist shop}};
        \end{tikzpicture}&
        \adjustbox{raise =5.45 ex}
         {
         \begin{tikzpicture}
            \node[inner sep=0pt] (russell) at (0,0) {\includegraphics[width=0.12\textwidth]{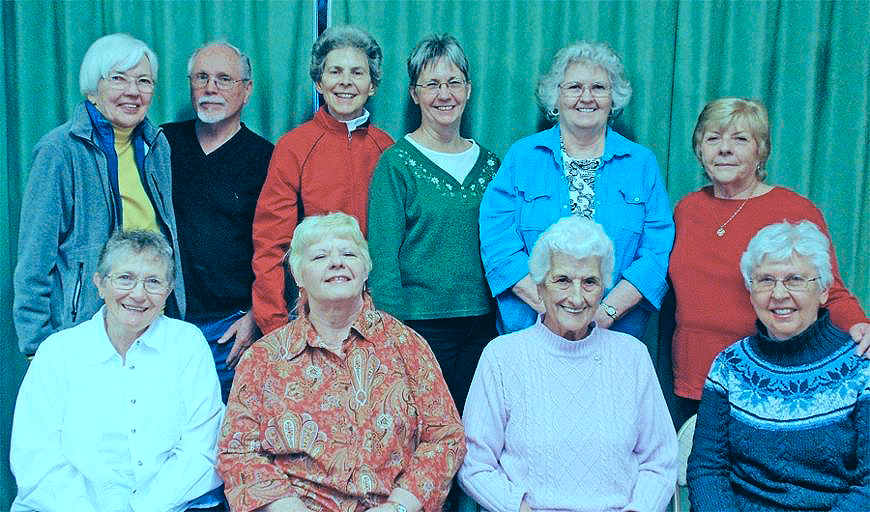}};
            \draw[fill=black] (-0.5pt,-11pt) rectangle (29.5pt,-18pt) node[pos=.5] {\tiny \textcolor{white}{medina}};
        \end{tikzpicture}
        }&
         \adjustbox{raise =5.45 ex}
         {
         \begin{tikzpicture}
            \node[inner sep=0pt] (russell) at (0,0) {\includegraphics[width=0.12\textwidth]{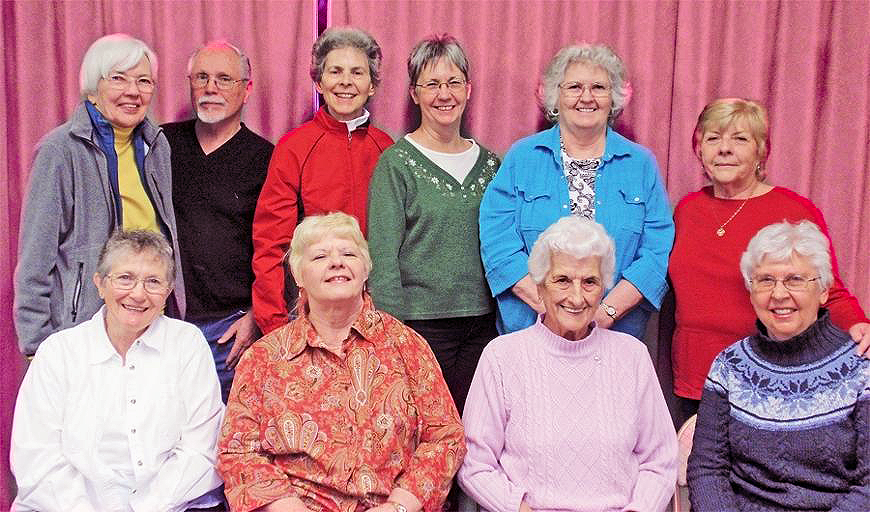}};
            \draw[fill=black] (-0.5pt,-11pt) rectangle (29.5pt,-18pt) node[pos=.5] {\tiny \textcolor{white}{throne room}};
        \end{tikzpicture}
         }\\
        \begin{tikzpicture}
            \node[inner sep=0pt] (russell) at (0,0) {\includegraphics[width=0.12\textwidth]{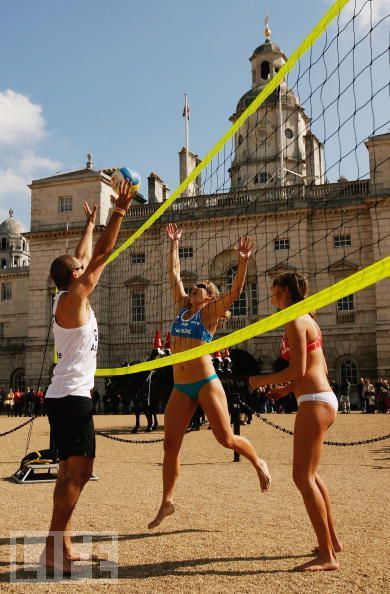}};
            \draw[fill=black] (-0.5pt,-38pt) rectangle (29.5pt,-45pt) node[pos=.5] {\tiny \textcolor{white}{volleyball}};
        \end{tikzpicture}&
        \adjustbox{raise =6.9 ex}
        {
        \begin{tikzpicture}
            \node[inner sep=0pt] (russell) at (0,0) {\includegraphics[width=0.12\textwidth]{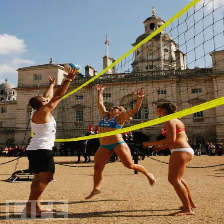}};
            \draw[fill=black] (-0.5pt,-23pt) rectangle (29.5pt,-30pt) node[pos=.5] {\tiny \textcolor{white}{horse cart}};
        \end{tikzpicture}
        }&
         \adjustbox{raise =6.9 ex }
         {
         \begin{tikzpicture}
            \node[inner sep=0pt] (russell) at (0,0) {\includegraphics[width=0.12\textwidth]{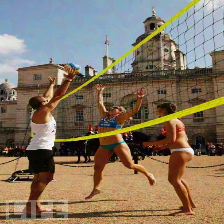}};
            \draw[fill=black] (-0.5pt,-23pt) rectangle (29.5pt,-30pt) node[pos=.5] {\tiny \textcolor{white}{horse cart}};
        \end{tikzpicture}
         }&
         \adjustbox{raise =6.9 ex}{
         \begin{tikzpicture}
            \node[inner sep=0pt] (russell) at (0,0) {\includegraphics[width=0.12\textwidth]{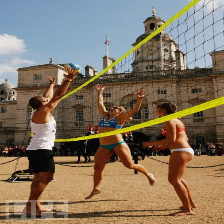}};
            \draw[fill=black] (-0.5pt,-23pt) rectangle (29.5pt,-30pt) node[pos=.5] {\tiny \textcolor{white}{horse cart}};
        \end{tikzpicture}
         }&
         \adjustbox{raise =6.9 ex}{
         \begin{tikzpicture}
            \node[inner sep=0pt] (russell) at (0,0) {\includegraphics[width=0.12\textwidth]{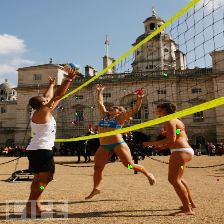}};
            \draw[fill=black] (-0.5pt,-23pt) rectangle (29.5pt,-30pt) node[pos=.5] {\tiny \textcolor{white}{bubble}};
        \end{tikzpicture}
         }&
         \adjustbox{raise =6.9 ex}{
         \begin{tikzpicture}
            \node[inner sep=0pt] (russell) at (0,0) {\includegraphics[width=0.12\textwidth]{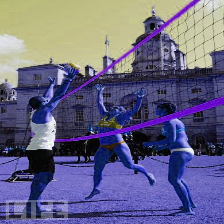}};
            \draw[fill=black] (-0.5pt,-23pt) rectangle (29.5pt,-30pt) node[pos=.5] {\tiny \textcolor{white}{bubble}};
        \end{tikzpicture}
         }&
         \begin{tikzpicture}
            \node[inner sep=0pt] (russell) at (0,0) {\includegraphics[width=0.12\textwidth]{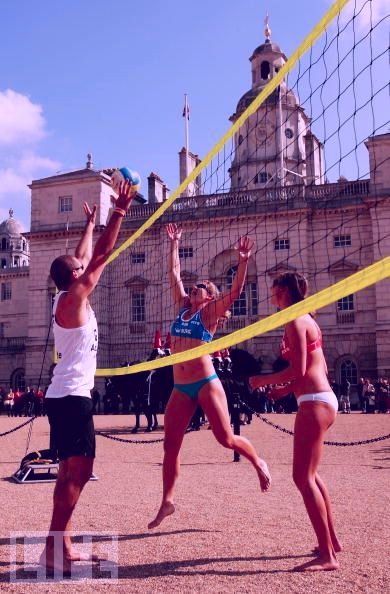}};
            \draw[fill=black] (-0.5pt,-38pt) rectangle (29.5pt,-45pt) node[pos=.5] {\tiny \textcolor{white}{maypole}};
        \end{tikzpicture}&
         \begin{tikzpicture}
            \node[inner sep=0pt] (russell) at (0,0) {\includegraphics[width=0.12\textwidth]{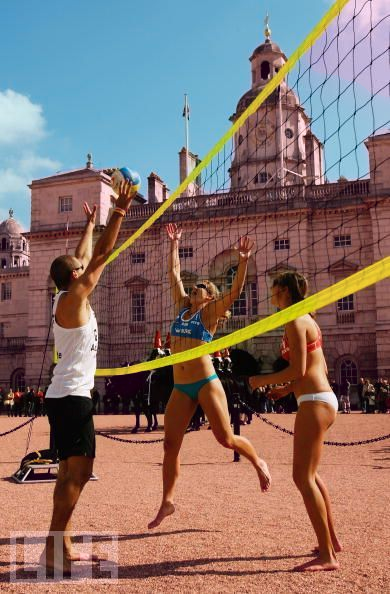}};
            \draw[fill=black] (-0.5pt,-38pt) rectangle (29.5pt,-45pt) node[pos=.5] {\tiny \textcolor{white}{horse cart}};
        \end{tikzpicture}\\
    \end{tabular}
    }
      \caption{Adversarial image samples from Private-Places365 (first row) and ImageNet (second row) datasets generated by Basic Iterative Method (BIM), Translation-Invariant BIM (TI-BIM), DeepFool (DF), SparseFool (SF), SemanticAdv (SA), ColorFool-r (CF-r) and the proposed ColorFool (CF). Please note that CF-r and CF generate examples at the native image resolution. The predicted class is shown on the bottom right of each image.}
      \label{fig:advExamples}
\end{figure*}
Another observation is that the robustness of the adversarial images is proportional to the accuracy of the classifier used for their generation (see Figs.~\ref{fig:robustness},~\ref{fig:detectability}). For example, misleading a highly-accurate DNN such as R50, which obtains an accuracy of almost $0.95$ on CIFAR-10, needs larger perturbations, which increase the robustness but also the detectability.

Tab.~\ref{tab:R-Adv-PCL} shows the SR of adversarial attacks in misleading ResNet110~\cite{he2016deep} trained on CIFAR-10 as well as the robustness to an improved training procedure based on PCL~\cite{Mustafa_2019_ICCV} and its combination with adversarial training~\cite{goodfellow2014explaining}. For the adversarial training, ResNet110 is trained on the clean and adversarial images generated by BIM (i.e.~the strongest defense~\cite{he2019parametric}). Tab.~\ref{tab:R-Adv-PCL} shows that ColorFool is robust as its SR remains above 99\% when misleading ResNet110 equipped with both PCL and adversarial training defenses. Instead, the SR of restricted adversarial attacks drops considerably.

\myparagraph{Quality.}
Sample adversarial images are shown in Fig.~\ref{fig:advExamples}. For instance, even though the restricted attacks such as TI-BIM or SparseFool generate adversarial images with minimal perturbations, they are noticeable. SemanticAdv and ColorFool-r generate unrealistic colors. However, even if ColorFool generates adversarial images that are largely different (in an $L_p$-norm sense) from the clean images, they look natural. Moreover, ColorFool generates images with the same dimensions as the clean images. The results of the image quality evaluation are shown in Tab.~\ref{tab:Q-PC}.
Unrestricted attacks obtain the highest NIMA scores across all attacks, classifiers and datasets. Specifically, in P-Places365 and ImageNet, ColorFool-r and ColorFool obtain the highest scores (over 5.19). For CIFAR-10, SemanticAdv, ColorFool-r and ColorFool obtain similar results with scores over 4.96. This implies that adversarial images generated by ColorFool do not deteriorate the perceived image quality while restricted attacks such as DeepFool or SparseFool obtain slightly lower results.
ColorFool obtains equal or higher NIMA scores than the clean images considering all datasets and classifiers.

\begin{table}[t]
    \centering
    \setlength{\tabcolsep}{1.5pt}
    \caption{Image quality (NIMA, the higher the better) of adversarial images from the Private-Places365 dataset, CIFAR-10 and ImageNet datasets for all adversarial attacks against ResNet50 (R50), ResNet18 (R18) and AlexNet (AN). We report only the mean value as the standard deviations are similar across all attacks with typical values of 4.4. KEY -- AC: attacked classifier. In bold the best performing attack per classifier and dataset.}
    \vspace{0.2cm}
    \begin{tabular}{l|rrr|rrr|rrr}
    \Xhline{3\arrayrulewidth} 
        {Dataset} & \multicolumn{3}{c|}{{P-Places365}} & \multicolumn{3}{c|}{{CIFAR-10}} & \multicolumn{3}{c}{{ImageNet}}  \\ \hline 
        \diagbox[width=4em]{{Attack}}{{AC}} & \multicolumn{1}{c}{R50} &  \multicolumn{1}{c}{R18} &  \multicolumn{1}{c|}{AN} &  \multicolumn{1}{c}{R50} &  \multicolumn{1}{c}{R18} &  \multicolumn{1}{c|}{AN} &  \multicolumn{1}{c}{R50} &  \multicolumn{1}{c}{R18} &  \multicolumn{1}{c}{AN} \\
        \Xhline{3\arrayrulewidth} 
        Clean & 5.02  & 5.02  & 5.02  & 4.91  & 4.91  & 4.91  & 5.23  & 5.23  & 5.23   \\ \hline 
        BIM & 4.88  & 4.88  & 4.85  & 4.90  & 4.90  & 4.92  & 4.88  & 4.89  & 4.87  \\
        TI-BIM & 4.92 & 4.92  & 4.86  & 4.92  & 4.92  & 4.95 & 4.83 & 4.83 & 4.77   \\
        DF & 4.95  & 4.94  & 4.94  & 4.88  & 4.88  & 4.92  & 4.93  & 4.93  & 4.92  \\
        SF & 4.99  & 4.99 & 4.97  & 4.86  & 4.86 & 4.87  & 4.97  & 4.96  & 4.94   \\
        SA & 5.05  & 5.05  & 5.06  & 5.01  & 5.01  & \textbf{4.98}  & 4.80  & 4.79  & 4.80  \\
        CF-r & \textbf{5.24} & \textbf{5.22} & \textbf{5.20} & \textbf{5.05} & \textbf{5.04} & 4.96 & \textbf{5.24}  & \textbf{5.25} & \textbf{5.23} \\
        CF & 5.22  & \textbf{5.22} & 5.19  & 5.04 & 5.03 & 4.96 & \textbf{5.24} & 5.24 & \textbf{5.23} \\
    \Xhline{3\arrayrulewidth} 
    \end{tabular}
    \label{tab:Q-PC}
\end{table}

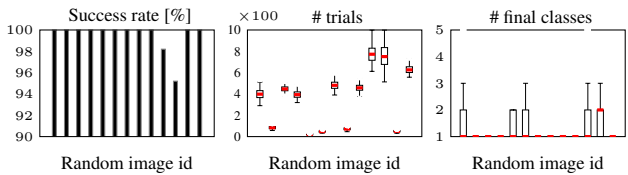
\begin{figure}[t!]
\pgfplotsset{
    every non boxed x axis/.style={},
    boxplot/every box/.style={solid,ultra thin,black},
    boxplot/every whisker/.style={solid,ultra thin,black},
    boxplot/every median/.style={solid,very thick, red},
}
\pgfplotstableread{numIters.txt}\numiters
\pgfplotstableread{finalClasses.txt}\finalClasses
\pgfplotstableread[row sep=\\,col sep=&]{
    id    & sr \\
    1     & 100 \\
    2     & 100 \\
    3     & 100 \\
    4     & 100 \\
    5     & 100 \\
    6     & 100 \\
    7     & 100 \\
    8     & 100 \\
    9     & 100 \\
    10     & 98.2 \\
    11     & 95.2 \\
    12     & 100 \\
    13     & 100 \\
    }\sr
    \begin{tikzpicture}
    \node[inner sep=0pt] (whitehead) at (1.2,1.60)
         {\scriptsize Success rate [\%]};
    \scriptsize
    \begin{axis}[
        width=3.9cm,
        height=3cm,
        bar width=1.5pt,
        ymin=90,  ymax=100,
        ytick={90,92,94,96,98,100},
        xtick=data,
        xticklabels={,,,},
        y label style={font=\scriptsize, at={(axis description cs:.5,.5)},anchor=south},
        label style={font=\tiny},
        tick label style={font=\tiny},
        xtick style={draw=none},
        ytick style={draw=none},
        xlabel={\scriptsize Random image id}
    ]
    \addplot[ybar=0pt, black, fill=black, draw opacity=0.5] table[x=id,y=sr]{\sr};
    \end{axis}
    \end{tikzpicture}
    \begin{tikzpicture}
    \scriptsize
             \node[inner sep=0pt] (whitehead) at (.14,1.60)
          {\tiny $\times 100$};
        \node[inner sep=0pt] (whitehead) at (1.2,1.60)
         {\scriptsize \# trials};
        \begin{axis}[
            width=3.9cm,
            height=3cm,
            xmin=0, xmax=14,
            boxplot/draw direction=y,
            xtick={1,...,13},
            xticklabels={,,,},
            ymin=0,ymax=1000,
            ytick={0,200,400,600,800,1000},
            yticklabels={0,2,4,6,8,10},
            label style={font=\tiny},
            y label style={font=\tiny, at={(axis description cs:.5,.5)},anchor=south},
            tick label style={font=\tiny},
            xtick style={draw=none},
            ytick style={draw=none},
            xlabel={\scriptsize Random image id}
        ]
        \addplot+[boxplot, boxplot/draw position=1,mark=*, mark options={white,scale=0.5},boxplot/box extend=0.5] table[y=A]{\numiters};
        \addplot+[boxplot, boxplot/draw position=2,mark=*, mark options={white,scale=0.5},boxplot/box extend=0.5] table[y=B]{\numiters};
        \addplot+[boxplot, boxplot/draw position=3,mark=*, mark options={white,scale=0.5},boxplot/box extend=0.5] table[y=C]{\numiters};
        \addplot+[boxplot, boxplot/draw position=4,mark=*, mark options={white,scale=0.5},boxplot/box extend=0.5] table[y=D]{\numiters};
        \addplot+[boxplot, boxplot/draw position=5,mark=*, mark options={white,scale=0.5},boxplot/box extend=0.5] table[y=E]{\numiters};
        \addplot+[boxplot, boxplot/draw position=6,mark=*, mark options={white,scale=0.5},boxplot/box extend=0.5] table[y=F]{\numiters};
        \addplot+[boxplot, boxplot/draw position=7,mark=*, mark options={white,scale=0.5},boxplot/box extend=0.5] table[y=G]{\numiters};
        \addplot+[boxplot, boxplot/draw position=8,mark=*, mark options={white,scale=0.5},boxplot/box extend=0.5] table[y=H]{\numiters};
        \addplot+[boxplot, boxplot/draw position=9,mark=*, mark options={white,scale=0.5},boxplot/box extend=0.5] table[y=I]{\numiters};
        \addplot+[boxplot, boxplot/draw position=10,mark=*, mark options={white,scale=0.5},boxplot/box extend=0.5] table[y=J]{\numiters};
        \addplot+[boxplot, boxplot/draw position=11,mark=*, mark options={white,scale=0.5},boxplot/box extend=0.5] table[y=K]{\numiters};
        \addplot+[boxplot, boxplot/draw position=12,mark=*, mark options={white,scale=0.5},boxplot/box extend=0.5] table[y=L]{\numiters};
        \addplot+[boxplot, boxplot/draw position=13,mark=*, mark options={white,scale=0.5},boxplot/box extend=0.5] table[y=M]{\numiters};
        \end{axis}
        \end{tikzpicture}
        \begin{tikzpicture}
        \node[inner sep=0pt] (whitehead) at (1.2,1.60)
         {\scriptsize \# final classes};
        \scriptsize
        \begin{axis}[
            width=3.9cm,
            height=3cm,
            xmin=0, xmax=14,
            boxplot/draw direction=y,
            xtick={1,...,13},
            ymin=1,ymax=5,
            xticklabels={,,,},
            ytick={1,2,3,4,5},
            label style={font=\tiny},
            tick label style={font=\tiny},
            xtick style={draw=none},
            ytick style={draw=none},
            xlabel={\scriptsize Random image id}
        ]
        \addplot+[boxplot, boxplot/draw position=1,mark=*, mark options={white,scale=0.5},boxplot/box extend=0.5] table[y=A]{\finalClasses};
        \addplot+[boxplot, boxplot/draw position=2,mark=*, mark options={white,scale=0.5},boxplot/box extend=0.5] table[y=B]{\finalClasses};
        \addplot+[boxplot, boxplot/draw position=3,mark=*, mark options={white,scale=0.5},boxplot/box extend=0.5] table[y=C]{\finalClasses};
        \addplot+[boxplot, boxplot/draw position=4,mark=*, mark options={white,scale=0.5},boxplot/box extend=0.5] table[y=D]{\finalClasses};
        \addplot+[boxplot, boxplot/draw position=5,mark=*, mark options={white,scale=0.5},boxplot/box extend=0.5] table[y=E]{\finalClasses};
        \addplot+[boxplot, boxplot/draw position=6,mark=*, mark options={white,scale=0.5},boxplot/box extend=0.5] table[y=F]{\finalClasses};
        \addplot+[boxplot, boxplot/draw position=7,mark=*, mark options={white,scale=0.5},boxplot/box extend=0.5] table[y=G]{\finalClasses};
        \addplot+[boxplot, boxplot/draw position=8,mark=*, mark options={white,scale=0.5},boxplot/box extend=0.5] table[y=H]{\finalClasses};
        \addplot+[boxplot, boxplot/draw position=9,mark=*, mark options={white,scale=0.5},boxplot/box extend=0.5] table[y=I]{\finalClasses};
        \addplot+[boxplot, boxplot/draw position=10,mark=*, mark options={white,scale=0.5},boxplot/box extend=0.5] table[y=J]{\finalClasses};
        \addplot+[boxplot, boxplot/draw position=11,mark=*, mark options={white,scale=0.5},boxplot/box extend=0.5] table[y=K]{\finalClasses};
        \addplot+[boxplot, boxplot/draw position=12,mark=*, mark options={white,scale=0.5},boxplot/box extend=0.5] table[y=L]{\finalClasses};
        \addplot+[boxplot, boxplot/draw position=13,mark=*, mark options={white,scale=0.5},boxplot/box extend=0.5] table[y=M]{\finalClasses};
        \end{axis}
        \end{tikzpicture}
    \caption{Influence of the randomness in the generation of ColorFool adversarial images on the success rate, the number of trials to converge and the number of different final classes at convergence with thirteen random images (500 random initializations) from ImageNet when attacking ResNet50.}
    \label{fig:randomness}
\end{figure}

\myparagraph{Randomness analysis.} 
As ColorFool generates random perturbations, we analyze what is the effect of this randomness on the SR, the number of trials to converge and whether the predicted class of the generated adversarial image varies. We execute ColorFool 500 times with thirteen random images from ImageNet that belong to different classes for attacking R50. We select R50 for this analysis as it is the most accurate classifier among the considered ones. 
Fig.~\ref{fig:randomness} shows the SR, the statistics (median, min, max, 25 percentile and 75 percentile) of the number of trials to converge and the number of classes that the executions converge to. Results for different images are shown on the x-axis.
We can observe that the number of trials that ColorFool requires to converge remains with a low median and standard deviation for images that always succeed in misleading the classifier (see the first and second plot in Fig.~\ref{fig:randomness}). 
Finally, most of the executions for a given image converge to the same class (see median value in the third plot in Fig.~\ref{fig:randomness}), regardless of the randomness.

\section{Conclusion}
\label{sec:conclusion}
We proposed a novel black-box adversarial attack, ColorFool, that modifies the color of semantic regions in an image based on priors on color perception. 
ColorFool achieves state-of-the-art results regarding success rate in misleading seen and unseen classifiers, robustness to defenses that employ filters, adversarial training or improved training loss function, as well as being less detectable than restricted attacks, especially JPEG compression. Furthermore, ColorFool generates adversarial images with the same size as the clean images.
We hope that our work will encourage studies on adversarial attacks that simultaneously consider the human visual system and the semantic information of the objects in the image, and new defenses against colorization to make DNNs robust to color changes.

As future work, we will evaluate adversarial attacks under a larger set of defenses and explore the behavior of adversarial attacks based on colorization in tasks such as object detection and semantic segmentation.

\section*{Acknowledgements}
This work is supported in part by the CHIST-ERA program through the project CORSMAL, under UK EPSRC grant EP/S031715/1.
Andrea Cavallaro wishes to thank the Alan Turing Institute (EP/N510129/1), which is funded by the EPSRC, for its support through the project PRIMULA.

\end{document}